\pdfoutput=1

\documentclass[11pt]{article}

\usepackage[final]{emnlp2021}

\usepackage{times}
\usepackage{latexsym}

\usepackage[T1]{fontenc}

\usepackage[utf8]{inputenc}

\usepackage{microtype}
\usepackage{xspace}

\usepackage{todonotes}

\newcommand{\ours}{\textsc{Dialki}\xspace} 

\usepackage[utf8]{inputenc}
\usepackage{graphicx}
\usepackage{amssymb}
\usepackage{amsmath}
\usepackage{esvect}
\usepackage{svg}
\usepackage{comment}
\newcommand\norm[1]{\left\lVert#1\right\rVert}

\usepackage{comment}
\usepackage{subcaption}
\usepackage{graphicx}

\usepackage{booktabs, multirow} 
\usepackage{soul}
\usepackage{makecell}
\usepackage{threeparttable}
\usepackage{footnote}
\makesavenoteenv{tabular}

\title{\ours: Knowledge Identification in Conversational Systems through Dialogue-Document Contextualization 
}

\newcommand\ai{$^{\diamondsuit}$}
\newcommand\uw{$^\spadesuit$}
\newcommand\first{$^\ast$}
\newcommand\aspace{\hspace{.75em}}

\author{
    Zeqiu Wu \uw\thanks{\first Equal contribution}\aspace
    Bo-Ru Lu \uw\first\aspace
    Hannaneh Hajishirzi \uw\ai\aspace
    Mari Ostendorf \uw\aspace\\
    \uw University of Washington \aspace
    \ai Allen Institute for AI\\
    {\tt \{zeqiuwu1,roylu,hannaneh,ostendor\}@washington.edu}
}

\begin{document}
\maketitle
\begin{abstract}
Identifying relevant knowledge to be used in conversational systems that are grounded in long documents is critical to effective response generation.
We introduce a knowledge identification model that leverages the document structure to provide dialogue-contextualized passage encodings and better locate knowledge relevant to the conversation. An auxiliary loss captures the history of dialogue-document connections.
We demonstrate the effectiveness of our model on two document-grounded conversational datasets
and provide analyses showing generalization to unseen documents and long dialogue contexts. 
\end{abstract}

\section{Introduction}
\label{sec:intro}


Many conversational agent scenarios require knowledge-grounded response generation, where knowledge is represented in written documents.  Most prior work has explored architectures for knowledge grounding in an end-to-end framework, optimizing a loss function targeting response generation \cite{Ghazvininejad2018AKN,zhou2018dataset,Yavuz2018DeepCopy}.
However, the knowledge needed at any one turn in the dialogue is typically localized in the document, and some studies have shown that directly optimizing for knowledge extraction helps resolve complex user queries \cite{feng-etal-2020-doc2dial} and increases user engagement in social chat \cite{dinan2018wizard, Moghe2018TowardsEB}. 
For long documents, explicit knowledge identification can also be useful for model interpretability and human-in-the-loop assistant scenarios.

\begin{figure}[htp]
\centering
    \includegraphics[width=1\columnwidth]{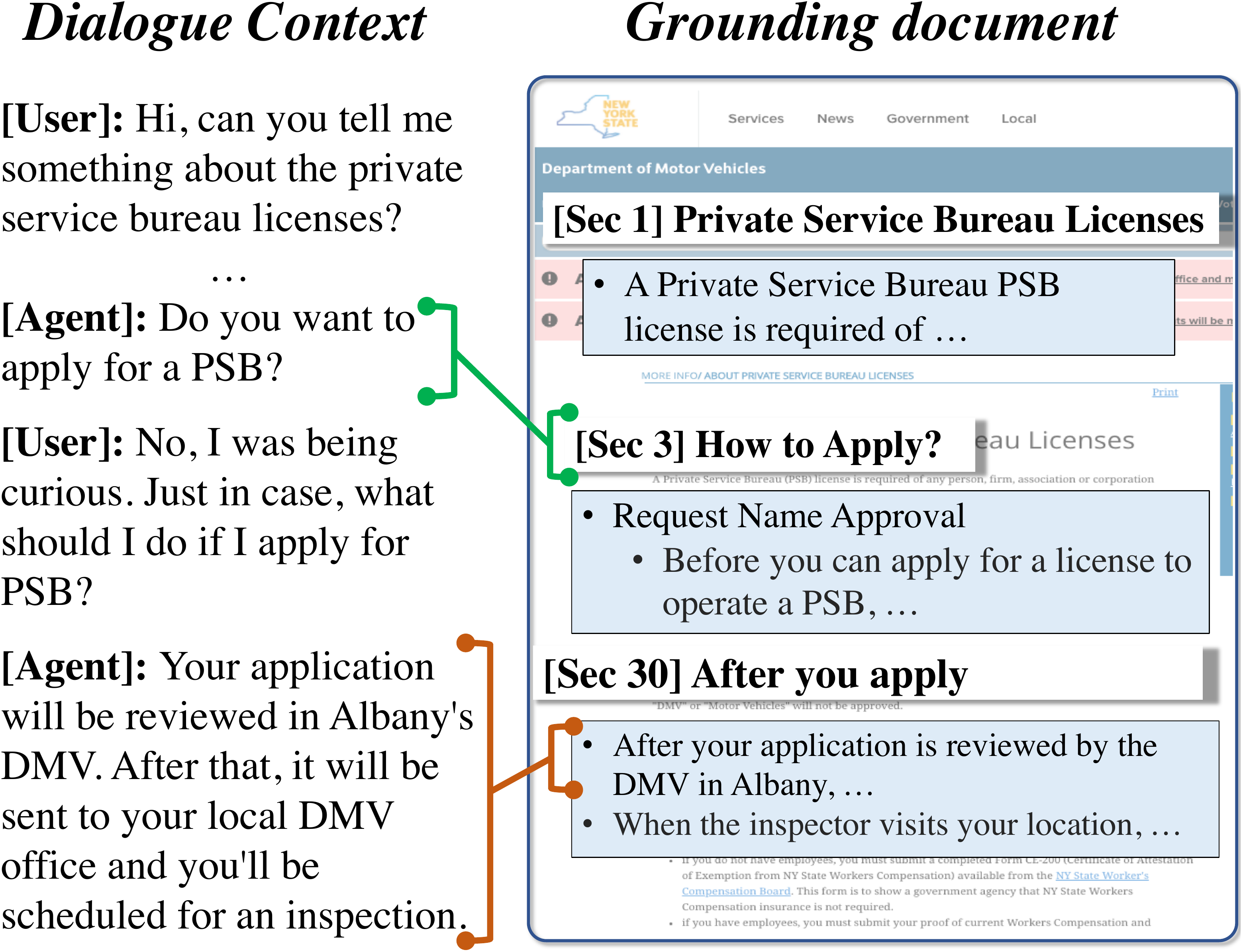}
    \caption{In a document-grounded conversation, \textit{knowledge identification} targets to locate a knowledge string within a long document to assist the agent in addressing the current user query. }
    \label{fig:teaser}
\end{figure}

Following~\cite{feng-etal-2020-doc2dial}, we define {\it knowledge identification} as the task of locating knowledge in a long document that is relevant to the current user query given the conversation context (Figure~\ref{fig:teaser}). 
Knowledge identification is similar to open question answering~\cite{chen-etal-2017-reading,min2020knowledge}, the task of answering a factoid question given a large grounding, except that it is not an interactive setting like dialogues.
With the assumption of a long grounding document, our task differs from prior work in conversational question answering \cite{choi-etal-2018-quac, reddy-etal-2019-coqa}, which focuses on answering a sequence of factoid questions about a short text snippet. Additionally, real user information needs can involve conversations with diverse forms of user queries and dialogue acts (e.g., asking for user preference, etc) as shown in Figure~\ref{fig:teaser}. 
Previous work in knowledge identification  encode the grounding document as a single string \cite{feng-etal-2020-doc2dial}, or splitting it into isolated sentences \cite{dinan2018wizard, Kim2020Sequential,zheng-etal-2020-difference} which potentially loses important discourse context.

In this paper, we introduce \ours to address knowledge identification in conversational systems with long grounding documents. In contrast to previous work, \ours extends multi-passage reader models in open question answering~\cite{karpukhin-etal-2020-dense, cheng-etal-2020-probabilistic} to obtain dense encodings of different spans in multiple passages in the grounding document, and it contextualizes them with the dialogue history. 
%
%
%
Specifically, \ours extracts knowledge given a long document by dividing it into paragraphs or sections and individually  contextualizes them with dialogue context. It then extracts knowledge by first selecting the most relevant passage to the dialogue context and then selecting the final knowledge string within the selected passage. Processing each passage rather than the full document greatly shortens the knowledge context, while preserving enough discourse context for reasoning. 
\ours also uses a multi-task objective to identify knowledge for the next turn, as well as used knowledge for previous turns that helps  improve the learning of both dialogue and document representations by capturing interdependencies between the next agent utterance, previous utterances, and the grounding document. 


Our model significantly improves over 
baselines on two conversational datasets, particularly on previously unseen documents or topics. Analyses show good generalization to longer grounding documents and longer dialogue context, as well as improvements in response generation. 
Model robustness can be further improved with an f-divergence based posterior regularization.  

Our contributions are summarized as follows. First, we propose a 
\textit{knowledge identification} model to address the problem of locating relevant information from long documents in a conversational context. Second, we introduce a 
multi-task learning framework that models the dialogue-document interactions via an auxiliary task of history knowledge prediction and a knowledge contextualization mechanism. Lastly, our model advances the state of the art in knowledge identification tasks for two conversational datasets, with more than 60\% and 20\% gains over previous work.\footnote{We release our source code for experiments at \url{https://github.com/ellenmellon/DIALKI}.} 
\section{Related Work}
\label{sec:related}

\paragraph{Conversational Question Answering.}
Existing conversational question answering tasks \cite{choi-etal-2018-quac, reddy-etal-2019-coqa} are generally defined as the task of reading a short text passage and answering a series of interconnected questions in a conversation. ShARC \cite{saeidi-etal-2018-interpretation} is a conversational machine reading dataset to address under-specified questions by requiring agents to ask follow-up questions grounded in a short text snippet that are answerable with ``yes/no'' answers.  These datasets focus on restricted dialogue act types (questions and answers) and short grounding texts. 
In contrast, our work focuses on more natural 
conversations that have a wider range of dialogue act types and are grounded in longer documents. 
In principle, models developed for short contexts could be applied to longer documents, but most have been based on pretrained language models (e.g., BERT \cite{devlin-etal-2019-bert} or RoBerTa \cite{liu2020roberta}) with an input that is the concatenation of a short document and dialogue context. Such language models usually have a limited maximum input length which constrains the implementation for long documents.

\paragraph{Knowledge-Grounded Dialogues.}
Most previous work in knowledge grounded dialogues \cite{Ghazvininejad2018AKN,zhou2018dataset, zhao-etal-2020-knowledge,lin-etal-2020-generating, li-etal-2019-incremental} generally focus on optimizing a loss function that targets response generation. Our work focuses on the knowledge identification task instead of response generation in a document-grounded dialogue. 

To the best of our knowledge, Wizard of Wikipedia (WoW) \cite{dinan2018wizard}, HollE \cite{Moghe2018TowardsEB} and Doc2Dial \cite{feng-etal-2020-doc2dial} are the only conversational datasets that include the task of knowledge identification in long documents. Both WoW and HollE focus on social chat conversations. WoW covers a much wider range of open-domain topics, while HollE is restricted to movie discussions. Doc2Dial is a more recent goal-oriented conversational dataset in various social welfare domains.

\paragraph{Knowledge Identification Models for Dialogues.}
Only a few works \cite{dinan2018wizard,lian-2019-postks,feng-etal-2020-doc2dial,Kim2020Sequential,zheng-etal-2020-difference} build models for and evaluate on the knowledge identification task. \citet{feng-etal-2020-doc2dial} encodes each long document as a single string without leveraging document structures, while \citet{dinan2018wizard,lian-2019-postks,Kim2020Sequential,zheng-etal-2020-difference} separately encode sentences in documents, which may have strong contextual dependencies among each other. Our model leverages document structures and divides each document into multiple passages to process. Similar to our model, \citet{Kim2020Sequential,zheng-etal-2020-difference} incorporate previously used knowledge, but they use a single vector to sequentially track the state of the used knowledge. Instead,  we apply a multi-task learning framework to model relations between grounding documents and history turns.

\section{Method}
\label{sec:method}

\begin{figure*}[htp]
\centering
    \includegraphics[width=\textwidth]{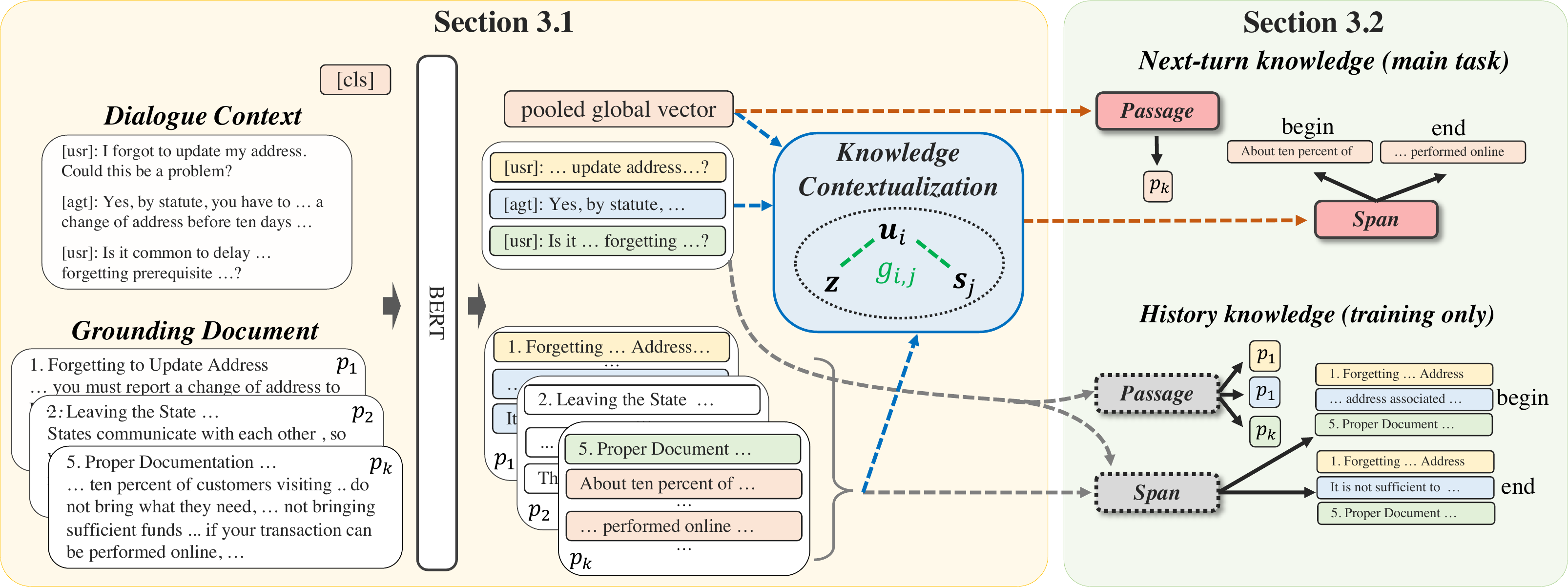}
    \caption{The overview of \ours. Each document is divided into passages. We apply BERT and a knowledge contextualization mechanism to obtain dialogue context and knowledge representations (left), for performing both next (main) and history (auxiliary) turn knowledge identification tasks (right). For each turn, \ours identifies knowledge by selecting the relevant passage as well as the begin/end spans in the passage.}
    \label{fig:model}
\end{figure*}

\paragraph{Problem Definition} 
\textit{Knowledge identification} in a document-grounded dialogue is defined as follows. At inference, given the dialogue context consisting of a sequence of $n$ previous utterances $(u_1,u_2,\dots,u_n)$ and a grounding document $\mathcal{D}$, select a substring $y$ in $\mathcal{D}$ that is relevant to the dialogue context and will be used in the next turn (i.e., utterance) in a conversation. We denote $u_1$ as the last (user) turn and $u_n$ as the earliest turn in dialogue history to help our model explanation.
 
Each document consists of a sequence of passages $\mathcal{D}=\{p_1, p_2, \dots, p_{|\mathcal{D}|}\}$ based on paragraphs or sections. Each passage $p$ consists of a sequence of semantic units $p=(s_1, s_2, \dots, s_{l})$, where each semantic unit (SU) can be either a token or a span or a sentence depending on how the document is segmented. For simplicity, we use ``span'' as the semantic unit in this section to describe our model. 

\paragraph{Method overview} In this section, we introduce \ours, a multi-task learning model for knowledge identification as illustrated in Figure~\ref{fig:model}.
We first introduce how we obtain dialogue utterance and knowledge span representations from BERT \cite{devlin-etal-2019-bert} and a span-level knowledge contextualization mechanism (Section~\ref{sec:encoding}).
These representations are then used for knowledge identification in our multi-task learning framework, which includes the main task of next-turn knowledge identification and an auxiliary task of history knowledge prediction applied during training only (Section~\ref{sec:multi-task}).
Finally, we describe our joint training objective and inference details (Section~\ref{sec:train_inference}).

\subsection{Encoding Dialog Context and Knowledge}
\label{sec:encoding}
\subsubsection{Multi-Passage Encoding}
\label{sec:multi-p-encoding}
Here, we describe how we initially obtain dense vector representations of each passage in the grounding document as a set of span representations. 
Inspired by the recent open-domain question answering multi-passage reader models \cite{karpukhin-etal-2020-dense, cheng-etal-2020-probabilistic}, we concatenate the dialogue context $(u_1,u_2,\dots,u_n)$ with $u_1$ to be the latest user turn, the document title $t$ and each passage $p$, and use a pretrained language model like BERT \cite{devlin-etal-2019-bert} to encode the concatenated sequence. More formally, the model input $\mathbf{X}$ for a passage $p$ of length $l$ becomes: 
\begin{eqnarray*}
 \mathbf{X} &= & [\text{cls}][\text{usr}] \, u_1 [\text{agt}] u_2 \cdots [\text{usr}] \, u_n 
    \\ &&  [\text{sep}]\, t \, [\text{cls}] \, s_1 \, [\text{cls}] \, s_2 \cdots [\text{cls}]\, s_{l} \,[\text{sep}]
\end{eqnarray*}
where `[usr]' and `[agt]' are special tokens indicating the start of a user or agent turn. `[cls]' indicates the start of the whole sequence or each knowledge span. `[sep]' are separator tokens. Then we
encode $\mathbf{X}$ and gather a sequence of pooled output vectors $\mathbf{H}=G(\text{BERT}(\mathbf{X}))$ where $G(.)$ gathers vectors of all `[cls]', `[usr]' and `[agt]' tokens. We decompose $\mathbf{H}$ as $[\mathbf{z}, \mathbf{u}_1,
\dots,\mathbf{u}_n, \mathbf{s}_1, \dots, \mathbf{s}_{l}]$ where $\mathbf{z}, \mathbf{u}_i, \mathbf{s}_j$ are pooled representations of the whole input (the first `[cls]' token in $\mathbf{X}$), dialogue utterance $u_i$ and span $s_j$ respectively. 


\subsubsection{Knowledge Contextualization}
\label{sec:know-ctx}
We leverage the pooled global, utterance and span representations $\mathbf{z},\mathbf{u}_i,\mathbf{s}_j$ obtained above for each passage $p$ to further contextualize knowledge span representations. Inspired by how EntNet \cite{henaff2016tracking} updates each entity representation based on an input sequence, for each span $s_j$ we calculate an updated span embedding $\mathbf{\widehat{s}}_j$ contextualized with the sequence of previous \textit{user} utterance representations as below. We use a gating function $g$ which determines how much the span embedding should be updated based on $\mathbf{u}_i$ and $\mathbf{z}$. 



\begin{equation*}
\begin{split}
    \mathbf{a}_{i,j} &= \mathbf{W}_s \, \mathbf{s}_j + \mathbf{W}_z \, \mathbf{z} + \mathbf{W}_u \, \mathbf{u}_i \, , \,  i \in \mathbf{C}_u\\
    \mathbf{\widehat{s}}_j &= \upsilon\Big(\sum\limits_{i\in \mathbf{C}_u} \Big[ \phi(\mathbf{a}_{i,j}) \odot g_{i,j}\Big] + \mathbf{s}_j\Big) \\
    g_{i,j} &= \sigma\Big({\mathbf{u}_i}^\top \, \mathbf{z} + {\mathbf{u}_i}^\top \, \mathbf{s}_j \Big),\\
\end{split}
\end{equation*}
where $\mathbf{W}_s,\mathbf{W}_z,\mathbf{W}_u \in \mathbb{R}^{d\times d}$ are model parameters, and 
$\mathbf{C}_u$ indexes the most recent user turns.
$\upsilon(.)$ is the vector normalization operation. $\sigma$ is the sigmoid function and $\phi$ can be any non-linear activation function. In our model, we use ReLU for $\phi$.  
Similarly, we calculate $\mathbf{\widetilde{s}}_j$ with previous \textit{agent} turns. Then new span embedding denotes as $\mathbf{\dot{s}}_j=[\mathbf{s}_j, \mathbf{\widehat{s}}_j, \mathbf{\widetilde{s}}_j]$. In our experiments, we set $\mathbf{C}_u$ to contain the most recent 2 user turns only, which leads to the best results.


\subsection{\ours with Multi-Task Learning }
\label{sec:multi-task}
\subsubsection{Next-Turn Knowledge Identification}
The main task in \ours is next-turn knowledge identification consisting of  two parts: passage prediction and knowledge span prediction (upper right in Figure~\ref{fig:model}).
At inference, given a set of passages $\{p_1, p_2, \dots, p_{|\mathcal{D}|}\}$ and the dialogue context $(u_1,u_2,\dots,u_n)$, our multi-passage knowledge identification model targets to predict the most relevant passage $p_k$ based on the
softmax probability in Eq. \eqref{eq:loss_p}, as well as the most relevant knowledge string $y=(s_b\dots s_e)$ in $p_k$ to the next agent turn in the conversation, based on the begin and end span probabilities in Eq. (\ref{eq:loss_b}-\ref{eq:loss_e}) 

We obtain the pooled global vector $\mathbf{z}$, each utterance and span representation $\mathbf{u}_i,\mathbf{\dot{s}}_j$ for each passage $p$ described in Section~\ref{sec:encoding}. We denote $\mathbf{Z}$ as the matrix containing the pooled global vectors for all passages and $\mathbf{U}_i$ as utterance representations for $u_i$ in all passages. $\mathbf{\dot{S}}$ denote the matrix with all span representations after knowledge contextualization.

\paragraph{Training Objectives} Eq. (\ref{eq:loss_p}-\ref{eq:loss_e}) show objective functions of knowledge passage $\mathcal{L}_{\text{psg}}$, begin $\mathcal{L}_{\text{begin}}$ and end $\mathcal{L}_{\text{end}}$ span predictions. $q(.)_t$ denotes the $t$-th index of the vector resulting from the softmax function. The variables $\hat{k}$, $\hat{b}$ and $\hat{e}$ correspond to the gold passage, begin and end span indices.
\begin{eqnarray}
    \mathcal{L}_{\text{psg}} &=& -\log \, q(\mathbf{W}_p \, \mathbf{Z})_{\hat{k}} \label{eq:loss_p} \\
    \mathcal{L}_{\text{begin}} &=& -\log \, q(\mathbf{W}_b \, \mathbf{\dot{S}})_{\hat{b}} \label{eq:loss_b} \\
    \mathcal{L}_{\text{end}} &=& -\log \, q(\mathbf{W}_e \, \mathbf{\dot{S}})_{\hat{e}} \label{eq:loss_e}
\end{eqnarray}
where $\mathbf{W}_p, \mathbf{W}_b, \mathbf{W}_e \in \mathbb{R}^d$. Therefore, the combined next turn knowledge selection loss function becomes:
\begin{equation*}
\mathcal{L}_{\text{next}} = \mathcal{L}_{\text{psg}} + \mathcal{L}_{\text{begin}} + \mathcal{L}_{\text{end}}.
\label{eq:loss_next}
\end{equation*}

\subsubsection{History Knowledge Identification}
\label{sec:hist-know}
The auxiliary task during training is to predict previously used knowledge, with the intuition that used knowledge in documents by history turns would guide the search for the next knowledge to use.
\paragraph{Training Objectives} With the same representations used in next-turn knowledge identification, similar to Eq. (\ref{eq:loss_p}-\ref{eq:loss_e}), we calculate both passage- and span-level prediction losses for each history turn $u_i$ if the knowledge string for $u_i$ can be found in $\mathcal{D}$. 
We calculate the passage prediction loss $\mathcal{L}^{h}_{\text{psg}}$ of previous turns as follows:
\begin{equation*}
    \mathcal{L}^{h}_{\text{psg}} = \frac{1}{\norm{\mathbf{U^*}}} \sum\limits_{u_i\in \mathbf{U^*}} -\log q\big(\mathbf{W}_p^h \,\,\phi(\mathbf{W}^h \, \mathbf{U}_i )\big)_{k_i} \label{eq:hist-p}
\end{equation*}
where $\mathbf{W}^h\in \mathbb{R}^{d\times d}, \mathbf{W}_p^h \in \mathbb{R}^{d}$ are model parameters. $\mathbf{U}^*$ is the set of history turns that can find their knowldge strings in the document $\mathcal{D}$. $k_i$ is the gold passage index for turn $u_i$. $\phi$ is a non-linear activation function, with ReLU used in our model.

Similarly, we calculate the average losses of predicting the gold begin $\mathcal{L}_{\text{begin}}^{h}$ and end $\mathcal{L}_{\text{end}}^{h}$ knowledge spans of all history turns in $\mathbf{U^*}$ in their gold passages. To do so, for each $u_i$, we compute the dot product between a linearly transformed $\mathbf{u}_i$ and each span embedding $\mathbf{s}_j$ in $p_{k_i}$ and apply a cross-entropy loss for the begin or end span prediction.
Therefore, the combined knowledge selection loss function of history turns becomes:
\begin{equation*}
\mathcal{L}_{\text{hist}} = \mathcal{L}_{\text{psg}}^{h} + \mathcal{L}_{\text{begin}}^{h} + \mathcal{L}_{\text{end}}^{h}.
\end{equation*}

\subsection{Training and Inference} 
\label{sec:train_inference}
\paragraph{Posterior Regularization}
\label{sec:posterior}
During training, we incorporate a posterior regularization mechanism \cite{cheng2020posterior} to enhance the model's robustness to domain shift. Specifically, we add an additional adversarial training loss as below. $\text{Div}$ is some f-divergence.\footnote{We use Jensen-Shannon divergence in all experiments.} Let $x$ be the encoded $\mathbf{X}$ (defined in Section~\ref{sec:multi-p-encoding}) after the BERT word embedding layer, \ours outputs $f_{psg}(x)$, $f_{begin}(x)$ and $f_{end}(x)$ as the next turn passage, begin and end knowledge span logits (the results before the softmax function $q(.)$ in Eq. (\ref{eq:loss_p}-\ref{eq:loss_e})) respectively.
\begin{equation*}
    \mathcal{L}_{\text{adv}} = \max_{\norm{\epsilon}\le a} \, \sum\limits_{f\in \{f_{psg}, f_{begin}, f_{end}\}} \, \hspace{-4mm}\text{Div}\big(f(x) || f(x+\epsilon)\big)
\end{equation*}
The above loss function essentially regularizes the $g$-based
worst-case posterior difference between the clean and noisy input (with norm of the added noise no larger than some scalar $a$) using an inner loop to search for the most adversarial direction.
\paragraph{Joint Training Objective} Combining all the above components, our final model optimizes the joint objective $\mathcal{L}$:
\begin{equation}
    \mathcal{L} = \mathcal{L}_{\text{next}} + \alpha\mathcal{L}_{\text{hist}} + \beta\mathcal{L}_{\text{adv}}
\label{eq:joint_objective}
\end{equation}
where $\alpha$ and $\beta$ are tunable hyperparameters.
\paragraph{Inference} During inference, we perform next-turn knowledge prediction only. We first predict the passage with the highest probability and enumerate over all possible knowledge span sequences in the selected passage with a maximum length. Then we select the span sequence with the highest score to be the final knowledge string. The score of each span sequence is calculated as the sum of begin and end span probabilities.
\section{Experiment}
\label{sec:exp}

\subsection{Datasets and Evaluation Metrics}
\paragraph{Datasets} We use two datasets for our experiments: \textit{Doc2Dial} \cite{feng-etal-2020-doc2dial} and \textit{Wizard of Wikipedia (WoW)} \cite{dinan2018wizard}. Doc2Dial contains about 4.8k goal-oriented dialogues in 4 social-welfare domains, with an additional Covid-19 domain in the blind held-out test set. Each dialogue has an average of 14 turns, grounded on a long document with more than 1k tokens on average. Each user or agent turn is grounded in a sequence of knowledge spans as annotated in the dataset.\footnote{Note that the knowledge identification task only targets to predict grounded knowledge for agent turns.} WoW contains over 20k social chat conversations with an average of 9 turns on over 1k open-domain topics. For each agent turn, the agent (i.e., wizard) chose one or no grounding sentence from on average 7 Wikipedia passages provided by a pre-defined retriever based on the dialogue history for composing the response. Each passage contains 10 sentences. The original data has its dev/test set split to two subsets, which contain conversations about topics seen or unseen in training.
\paragraph{Evaluation Metrics}
For evaluation, we use exact match (EM) and token-level F1 score as originally used in \citet{feng-etal-2020-doc2dial, dinan2018wizard}.

\subsection{Implementation Details}
\paragraph{Doc2Dial}
Documents are automatically split into passages by parsing the html files into smaller sections with different titles, resulting in an average of 8 passages per document. 
All passages from a single html file are used throughout each conversation.
Knowledge SUs are spans as segmented in the data. During inference, we set the maximum knowledge length to be 5 spans based on the dev set distribution.\footnote{We use data, baselines, preprocessing \& evaluation scripts at: \url{https://github.com/doc2dial/sharedtask-dialdoc2021}}
\paragraph{WoW}
Passage segmentation is automated in the WoW pre-defined retriever and the dataset provides 7 passages on average for each agent turn in the dataset. Since agents are allowed to select no grounding sentence, we add an additional passage with only one single sentence ``no passages used'' following the original data processing script\footnote{\url{https://github.com/facebookresearch/ParlAI}}. Knowledge SUs are sentences. We set the maximum knowledge length to be 1 sentence during inference. Passages may differ for each agent turn in the same conversation. Thus, during training, we only calculate history loss for previous agent turns whose ground truth knowledge can be found in next turn passages, which on average cover over 70\% of history agent turns.
\paragraph{Experimental Setup}
We initialize and finetune on BERT \cite{devlin-etal-2019-bert}. We use the uncased base BERT in most of our experiments, and set $3e^{-5}$ as the learning rates and 1000 as warm-up steps. For each experiment, we search the weights in Eq. \eqref{eq:joint_objective} on the dev set in the ranges of $\alpha=\{0.5, 1, 2\}$, $\beta=\{0.5, 2.5, 5\}$. We do not observe much difference with different weight combinations, but the best result is achieved when $\beta=5$ for both datasets, and $\alpha=1$ for Doc2Dial and $\alpha=0.5$ for WoW. Models are selected based on the best dev set EM score. The maximum length of dialogue context is 128. The maximum lengths of model input are 512 and 384 for Doc2Dial and WoW respectively. 
In training, we provide multiple passages from the grounding document as the input, where only one of them is the gold passage. We find that learning benefits from having more negative passage examples, and the number used is constrained by memory consumption (up to 10 for the models with posterior regularization and up to 20 otherwise).  
For inference, up to 20 passages from the target document are considered. For longer documents, the first 20 passages are used. 
Further details are in Appendix~\ref{sec:appendix_exp}.

\subsection{Compared Systems}
\paragraph{BERTQA-Token:} The original baseline \cite{feng-etal-2020-doc2dial} and the best published model on Doc2Dial. It uses BERTQA \cite{devlin-etal-2019-bert} with each dialogue context as the question and sliding windows to process each document, and predicts the start and end tokens in the document.
\paragraph{BERTQA-Span:} Similar to BERTQA-Token, but predicts the start and end knowledge spans instead of tokens. Instead of using sliding windows, we increase the number of position embeddings to be 2048, initialized with 512 position embeddings in BERT repeated 4 times, following \cite{beltagy2020longformer}. We observe better results with this operation than when using sliding windows.
\paragraph{Transformer MemNet:} The original baseline \cite{dinan2018wizard} of WoW, which uses  a vanilla Transformer \cite{vaswani2017attention} to encode all knowledge sentences separately and a memory network for sentence selection. Another model version includes pre-training on Reddit conversations.
\paragraph{SLKS:} The state-of-the-art model \cite{Kim2020Sequential} on WoW that encodes all knowledge sentences and dialogue turns separately with BERT (or RNN). It uses two GRUs to update the states of dialog history and previously selected sentences.
\paragraph{DiffKS:} This model \cite{zheng-etal-2020-difference} is similar to SLKS. Additionally, it computes the representation difference between each candidate knowledge sentence and the state of previously used knowledge for in the final decision function.  
\paragraph{Multi-Sentence:} This baseline is designed to be similar to \ours, but divides documents into sentences instead of passages. It calculates the next knowledge prediction loss $\mathcal{L}_{next}$ only. For Doc2Dial, we use subsections, mostly single sentences, as segmented in documents. Knowledge strings rarely exceed the subsection boundaries.
\paragraph{\ours (Ours):} Our multi-passage knowledge identification model with the next turn knowledge prediction loss $\mathcal{L}_{\text{next}}$, history knowledge prediction loss $\mathcal{L}_{\text{hist}}$, contextualization mechanism (\textit{know-ctx}) and posterior regularization loss $\mathcal{L}_{\text{adv}}$.

\subsection{Quantitative Results}
\paragraph{Doc2Dial} Table~\ref{tab:doc2dial_test} reports the results of different systems in the  blind held-out test set with an unseen Covid-19 domain. All models are based on the BERT-base model except the last one that uses BERT-large. 
The full model of \ours achieves best results, demonstrating the effectiveness of combining all components of the system described in Section~\ref{sec:method}.
The significant improvement from \ours ($\mathcal{L}_{\text{next}}$ only) 
over BERTQA-Token, which takes the full document as a single string, shows the benefit of our multi-passage framework. 
BERT-large helps further improve the overall results.\footnote{After ensemble with other large language models of RoBerTa and ELECTRA \cite{liu2020roberta, Clark2020ELECTRA}, our model achieves EM / F1 scores as 67.09 / 76.34, achieving the best scoring system on the leaderboard outperforming beating the second-best scores 63.53 / 75.94. }

\begin{table}
\centering
\small
\begin{tabular}{lrrr}\toprule
\multirow{2}{*}{\textbf{Method}}  &\multicolumn{2}{c}{\textbf{Overall}} \\
& EM &F1 \\\midrule
BERTQA-Token &34.6 &53.2 \\
BERTQA-Token (our version) &35.8 &52.6 \\
\ours ($\mathcal{L}_{\text{next}}$ only) &51.2 &64.7 \\
\ours &59.5 &71.0 \\ \hline
\ours (BERT-large) & \textbf{61.8} & \textbf{73.1} \\

\bottomrule
\end{tabular}
\caption{Evaluation results on the Doc2Dial test set.}
\label{tab:doc2dial_test}
\end{table}

\begin{table}
\centering
\resizebox{\columnwidth}{!}{
\begin{tabular}{lcccccc}\toprule
\multirow{2}{*}{\textbf{Method}} &\multicolumn{2}{c}{\textbf{Seen}} &\multicolumn{2}{c}{\textbf{Unseen}} \\
&EM &F1 &EM &F1 \\\midrule
Transformer MemNet &22.5 &33.2 &12.2 &19.8 \\
Transformer MemNet $+$ Pretrain &24.5 &36.4 &23.7 &35.8 \\
DiffKS (RNN) &25.5 &-- &19.7 &-- \\
SLKS (RNN) &23.4 &-- &14.7 &-- \\
SLKS (BERT-base) &26.8 &-- &18.3 &-- \\
Multi-Sentence (BERT-base) &30.4 &37.7 &27.6 &35.4 \\
\ours (BERT-base) &\textbf{32.9} &\textbf{40.7} &\textbf{35.5} &\textbf{43.4} \\
\bottomrule
\end{tabular}
}
\caption{Evaluation results of WoW test sets.}
\label{tab:wow_test}
\end{table}
\begin{table*}
\centering
\small
\resizebox{\textwidth}{!}{
\begin{tabular}{lcccccc|ccccccc}\toprule
\multirow{3}{*}{\textbf{Method}} &\multicolumn{6}{c}{\textbf{Doc2Dial}} &\multicolumn{6}{c}{\textbf{WoW}} \\
 &\multicolumn{2}{c}{\textbf{Overall}} &\multicolumn{2}{c}{\textbf{Seen}} &\multicolumn{2}{c}{\textbf{Unseen}} &\multicolumn{2}{c}{\textbf{Overall}} &\multicolumn{2}{c}{\textbf{Seen}} &\multicolumn{2}{c}{\textbf{Unseen}} \\
&EM &F1 &EM &F1 &EM &F1 &EM &F1 &EM &F1 &EM &F1 \\\midrule
BERTQA-Token &42.2 &58.1 &48.3 &61.1 &37.0 &55.6 &-- &-- &-- &-- &-- &-- \\
BERTQA-Span &46.3 &59.3 &54.4 &63.5 &39.4 &55.6 &-- &-- &-- &-- &-- &-- \\
Multi-Sentence &59.5 &68.8 &63.6 &71.6 &56.0 &66.4 &29.2 &37.0 &32.4 &39.7 &26.1 &34.3 \\\midrule
\ours ($\mathcal{L}_{\text{next}}$ only)  &60.4 &71.2 &64.2 &72.3 &57.1 &70.2 &31.5 &39.7 &33.3 &41.1 &29.8 &38.3 \\
\; $+ \mathcal{L}_{\text{hist}}$ &63.0 &72.6 &66.5 &73.9 &59.9 &71.9 &33.6 &41.6 &35.1 &42.7 &32.2 &40.5 \\
\; $+ \mathcal{L}_{\text{hist}},$ \textit{know-ctx} &63.8 &73.4 &\textbf{67.7} &74.8 &60.5 &72.3 &33.6 &41.5 &35.2 &42.8 &32.1 &40.3 \\
\; $+ \mathcal{L}_{\text{adv}}$ &64.4 &73.8 &66.2 &73.9 &62.8 &73.7 &32.9 &40.8 &34.6 &42.2 &31.1 &39.5 \\
\; $+ \mathcal{L}_{\text{hist}}, \mathcal{L}_{\text{adv}},$ \textit{know-ctx} &\textbf{65.9} &\textbf{74.8} &67.6 &\textbf{74.9} &\textbf{64.4} &\textbf{74.7} &\textbf{34.2} &\textbf{42.1} &\textbf{35.9} &\textbf{43.5} &\textbf{32.6} &\textbf{40.7} \\
\bottomrule
\end{tabular}
}
\caption{Ablation results on Doc2Dial and WoW dev sets.}
\label{tab:joint_ablation}
\end{table*}

\paragraph{WoW} Results on both test sets are presented in Table~\ref{tab:wow_test},  containing conversations on seen and unseen topics in training. \ours significantly outperforms all other systems, which encode knowledge sentences disjointly. This again confirms the advantage of our multi-passage framework and the modeling of dialogue-document relations.

Surprisingly, \ours achieves even higher results on the unseen test set while others observe performance drops. One potential reason is that blindly dividing passages into disjoint sentences to process may hurt the model's reasoning ability and generalization. In addition, Transformer MemNet, DiffKS and SLKS decouple the encoding of dialogue history and grounding sentences, which prevents the model to effectively reason over their relations. With more investigation into the data, we found that two thirds of the grounding passages used in those conversations about unseen topics actually appear in the training set based on title matching.\footnote{Only 1.6\% of agent turns in the WoW unseen (topic) test set have all of their retrieved passages not seen in training.} Moreover, all topics in the dataset are similar lifestyle topics collected from persona description sentences in \cite{zhang-etal-2018-personalizing}. These observations support the possibility of better performance on the original unseen set. 
\paragraph{Impact of different components of  \ours.}
Table~\ref{tab:joint_ablation} reports results of ablating different components of our system on the dev set of both datasets. Although Doc2Dial does not provide separate seen and unseen sets as WoW does, we split the dev set into examples that have grounding documents seen or unseen in the training set. Note that ``seen'' and ``unseen'' refer to documents and topics for Doc2Dial and WoW respectively. 

We observe that \ours consistently beat baseline models that either process the full document as a single string or isolated sentences. Our framework leads a to smaller performance gap between seen and unseen examples. Adding the auxiliary history knowledge prediction loss ($\mathcal{L}_{hist}$) leads to further improvements on both datasets. Adding \textit{know-ctx} helps enhance the performance on Doc2Dial while does not appear to be effective on WoW, as explored further below. 
Adding posterior regularization ($\mathcal{L}_{adv}$) is effective on both datasets while Doc2Dial gets more advantage from it especially on the unseen subset. Combining all model components yields the best results.

\subsection{Analysis}
\label{sec:analysis}

\paragraph{Impact on Response Generation}
\begin{table}
\centering
\resizebox{\columnwidth}{!}{
\begin{tabular}{llcc}\toprule
\textbf{KI Model} &\textbf{Knowledge Input} &\textbf{sacrebleu} \\\midrule
-- &full doc &22.84 \\
BERTQA-Token &pred span &21.42 \\
\ours &pred span &25.16 \\
\ours &pred span \& passage &\textbf{25.84} \\
\bottomrule
\end{tabular}
}
\caption{Response generation results on Doc2Dial dev set. KI stands for Knowledge Identification.}
\label{tab:doc2dial_generation}
\end{table} 

\begin{table*}
\centering
\small
\begin{tabular}{p{1.3cm}|p{6.9cm}|p{6.5cm}}
\toprule
Dialogue \newline Context & \textit{User}: I want to trade in my license for a New York one.  \newline
\textit{Agent}: You have to exchange your-out-state driver license within 30 days of \dots \newline
\textit{User}: What if I need my license for when I go back to my other country? \newline \dots \newline
\textit{Agent}: Are you studying in New York State? \newline
\textit{User}: No & 
\textit{User}: I heard something about co-op training program. Could you tell me about it, please? \newline
\textit{Agent}: A co-op training program refers to \dots \newline
\textit{User}: Sounds awesome. What should I do to get that?  \newline
\textit{Agent}: Well, for that, lets do a little fact check, shall we? Are you using VA educational assistance? \newline
\textit{User}: Yes, I'm using that. \\
\midrule
Baseline & Do you need a New York State driver license? & Good. In that case, you may be able to get money for books, tuition and housing. \\
Ours & Are you a driver from another country? & Are you enrolled at an approved institution of Higher Learning? \\
Gold \newline Response &
Are you a licensed driver from another country?  & Good. Now, are you enrolled at an approved institution of Higher Learning or IHL?\\
\bottomrule
\end{tabular}
\caption{\label{tab:gen_examples} Sample generated responses from BART with the full grounding document (baseline) or the predicted grounding span and passage by \ours (ours) as the additional input to the dialogue context.}
\vspace{-2mm}
\end{table*}
We apply BART \cite{lewis-etal-2020-bart} to decode agent responses given the concatenated dialogue context and grounding knowledge (e.g., document or predicted knowledge string) as the input. BART is also used as the baseline for the agent response generation task on Doc2Dial \cite{feng-etal-2020-doc2dial},\footnote{\url{https://github.com/doc2dial/sharedtask-dialdoc2021}} where the model is given the dialogue history concatenated with the full document to decode the next agent response. We conduct experiments on the same model architecture, with the knowledge input being the predicted knowledge string or passage. Without changing the model at all, using knowledge predicted by \ours leads to almost 3 points in the sacrebleu score \cite{post-2018-call}, as shown in
Table~\ref{tab:doc2dial_generation}. Examples of generated responses are shown in Table~\ref{tab:gen_examples}.

\paragraph{Passage Identification Accuracy}
We map predicted knowledge strings back to the passages and calculate the passage-level accuracy. Table~\ref{tab:passage_pred} shows that \ours outperforms baseline models in locating the passage containing the knowledge string. 
Notably, our models generalize well in passage prediction to unseen documents or dialogue topics. 

\begin{table}
\centering
\resizebox{\columnwidth}{!}{
\begin{tabular}{lcccccc}\toprule
\multirow{2}{*}{\textbf{Method}} &\multicolumn{2}{c}{\textbf{Doc2Dial}} &\multicolumn{2}{c}{\textbf{WoW}} \\
&Seen &Unseen &Seen &Unseen \\\midrule
BERTQA-Span &76.9 &72.7 &-- &-- \\
Multi-Sentence &85.3 &81.6 &68.0 &57.8 \\
\ours ($\mathcal{L}_{\text{next}}$ only) &86.6 &84.4  &72.9 &69.0 \\
\ours &\textbf{88.5} &\textbf{87.5} & \textbf{73.4} & \textbf{69.7} \\
\bottomrule
\end{tabular}
}
\caption{Passage prediction accuracy on dev sets.}
\label{tab:passage_pred}
\end{table}

\begin{figure}
  \centering
  \begin{subfigure}[b]{0.49\linewidth}
    \centering
    \includegraphics[width=\linewidth]{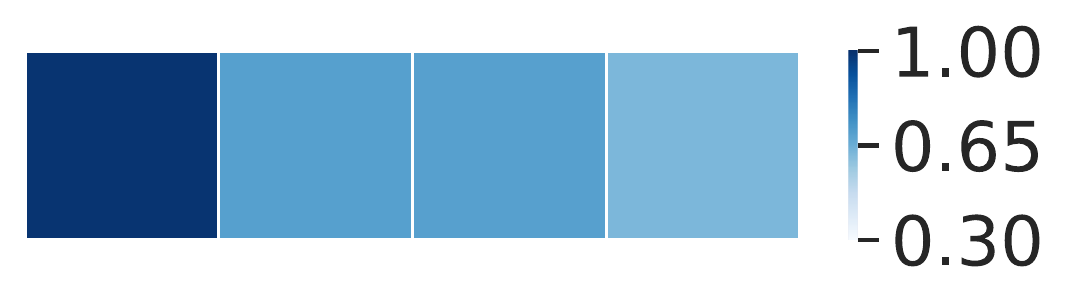}
    \label{subfig:doc2dial_att}
  \end{subfigure}
  \begin{subfigure}[b]{0.49\linewidth}
    \centering
    \includegraphics[width=\linewidth]{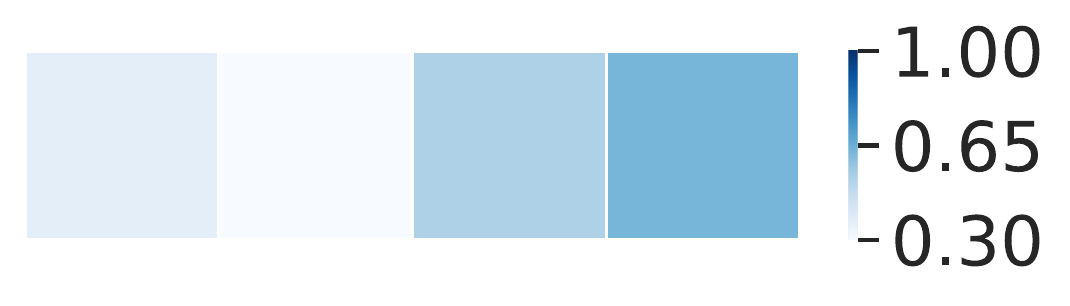}
    \label{subfig:wow_att}
  \end{subfigure}
\caption{Similarity between $\mathbf{z}$ and the latest 4 history turn representations  (i.e.,$\mathbf{u}_1\dots\mathbf{u}_4$ from left to right) on Doc2Dial (left) and WoW (right).}
\label{fig:att}
\end{figure}

\begin{figure*}
  \centering
  \includegraphics[width=1.0\textwidth]{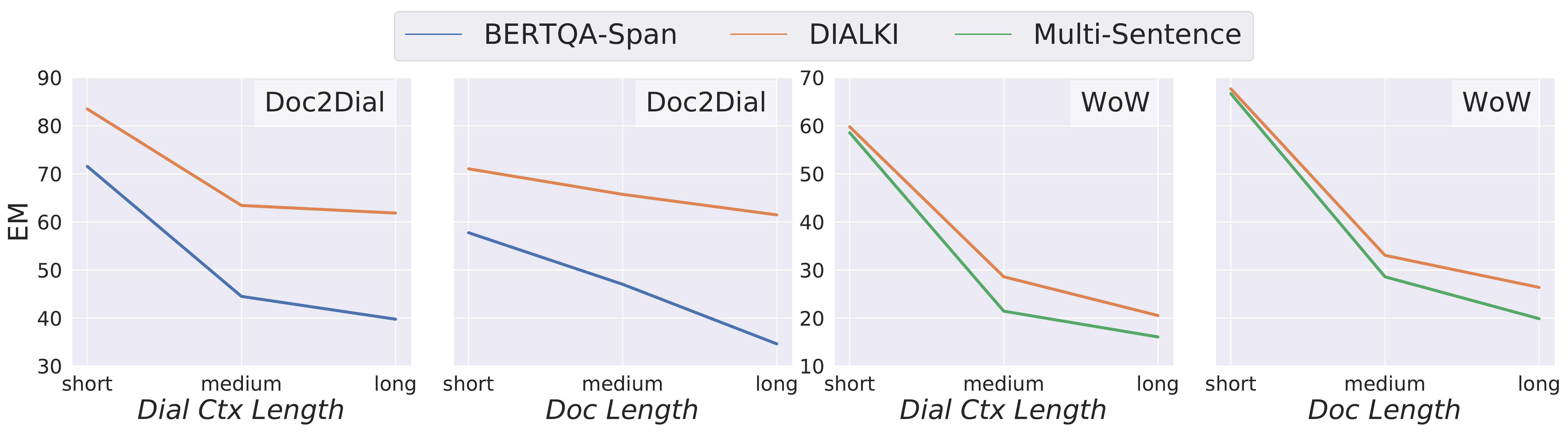}
  \caption{EM versus the length of dialogue context (\# previous turns) or document (\# tokens).}
 \label{fig:model_robustness}
\end{figure*}

\paragraph{Similarity Between Global and History Turn Representations}  
The dot product between $\mathbf{z}$ (the encoding of the whole input sequence) and each history utterance representation $\mathbf{u}_i$ (sigmoid normalized)
is used in the gating function $g$ (defined in Section~\ref{sec:know-ctx}) that gates the effect of each utterance $u_i$ in calculating span embeddings. Figure~\ref{fig:att} shows such normalized dot product scores between $\mathbf{z}$ and the latest four history turn vectors. 
In Doc2Dial, the score is relatively high for the more recent user turn and decreases for earlier turns. Such patterns are not observed in WoW. One potential reason is that each agent in a Doc2Dial conversation has a clear goal to directly address user queries, while WoW conversations are more like social chat. This distinction may explain why \textit{know-ctx} does not work well on WoW. The reason for even higher similarities with earlier turns in WoW could be that 
knowledge is related to people being referred to with pronouns with names being introduced earlier.

\paragraph{Availability of History Knowledge Labels}
In Doc2Dial, we calculate the history knowledge prediction loss ($\mathcal{L}_{hist}$) for all history turns, since all the labels are available in the training set. In practice, it might not be feasible to annotate all history turns, particularly user turns. 
Hence, we conduct experiments comparing
scenarios where history knowledge labels are given for all turns, agent turns only, random 50\% agent turns, or no turns. We get EM scores of 63.0, 62.7, 62.4 and 60.4 respectively, finding that removing user turn labels and half of agent turn labels do not affect the results much. 
\paragraph{Impact of Dialogue / Document Length}
Figure~\ref{fig:model_robustness} shows the average EM scores vs.\ the dialogue context and grounding document length
on the Doc2Dial and WoW dev set.
Dialogue context lengths are grouped into 0-2 (short), 3-5 (medium) and $\ge6$ (long) history turns. In Doc2Dial, documents are categorized as short, medium and long: 0-500, 501-1000 and 1000+ tokens, respectively. In WoW, documents are categorized as short, medium and long: 0-800, 801-1600 and 1600+ tokens, respectively.
\ours shows less performance drop as the two input lengths increase compared with baselines that do not leverage the multi-passage structure of grounding documents.
\paragraph{Span Prediction Error Analysis}
We randomly select and analyze 50 examples from both Doc2Dial and WoW where \ours makes wrong predictions (EM=0). Since \ours can achieve relatively high passage-level prediction accuracy as shown in Table~\ref{tab:passage_pred}, we focus on analyzing prediction errors where the passage is predicted correctly. Among the 50 examples from Doc2Dial, we find 6 error types of predicted knowledge: (1) overlapped with the ground truth knowledge string while including redundant or missing relevant details (38\%); (2) relevant to the user query but at an incorrect granularity level (18\%); (3) completely irrelevant (16\%); (4) relevant to history user queries instead of the current one (14\%); (5) contains keywords of the last user utterance that are irrelevant (10\%); (6) wrong gold labels (4\%). 

For WoW, the conversation style is different and the predicted knowledge is a single sentence instead of multiple spans, so prediction errors fall into different classes: (1) open-ended situations where the predicted knowledge is appropriate (48\%); (2) unnatural for use in the next response (30\%); (3) the predicted knowledge is more appropriate to use than the ground truth (12\%); (4) irrelevant knowledge that does not answer the user's questions (10\%). The high percentage of open-ended examples explains the relatively low evaluation scores of knowledge identification on WoW.

\section{Conclusion}

In summary, we introduce \ours to address  knowledge identification in conversational systems with long grounding documents, taking advantage of document structure to contextualize document passages together with the dialogue history. \ours uses a multi-task objective that identifies knowledge for the next turn and used knowledge for previous turns, which captures interconnections between the dialogue and the document.  Additional posterior regularization in learning further improves results. The model gives state of the art performance for this task 
on Doc2Dial and Wizard of Wikipedia, respectively.  We show that improvements in knowledge selection transfer to response generation with a baseline generator.

The current study is limited by the static nature of the available data. Further work is needed to assess performance in an interactive setting. It would also be of interest to consider scenarios where document information is irrelevant or users change their minds about what information they want.

\section*{Acknowledgments}
The research was supported by the ONR N00014-18-1-2826, DARPA N66001-19-2-4031, NSF (IIS 1616112), NSF CAREER award, and Allen Distinguished Investigator Award. We also thank members of UW-NLP who provided feedback and insights to this work. We would also like to thank Hao Cheng for providing insights for improving model robustness and thank Michael Lee for early discussions.

\section*{Ethical Considerations}
Our work is primarily intended to address and encourage future work on the task of knowledge identification in document-grounded dialogue systems. We believe that locating relevant knowledge to be used in the following conversation can be useful for improving the interpretability of response generation models. Knowledge identification can also play an important role in human-in-the-loop assistant scenarios. This places greater control into the hands of human agents instead of automatic response generation models, which tend to suffer from ethical issues like generating hallucinated \cite{zellers2019grover, wu2021a} or toxic content \cite{pavlopoulos-etal-2020-toxicity}.

\bibliography{emnlp2021}
\bibliographystyle{acl_natbib}

\newpage
\appendix
\section{Experimental Setup Details}
\label{sec:appendix_exp}
We initialize and finetune on BERT \cite{devlin-etal-2019-bert} downloaded from Huggingface Transformers \cite{wolf2019huggingface}.\footnote{\url{https://github.com/huggingface/transformers}} We use the uncased base model of BERT in most of our experiments, and set $3e^{-5}$ as the learning rates and 1000 as warm-up steps with linear decay. For each experiment, we search the weights in Eq. 4 on the dev set in the ranges of $\alpha=\{0.5, 1, 2\}$, $\beta=\{0.5, 2.5, 5\}$. We eventually use $\beta=5$ for all experiments, and $\alpha=1$ for Doc2Dial and $\alpha=0.5$ for WoW. We search for fewer than 5 hyperparameter trials for each experiment. All models are trained for 20 and 10 epochs for Doc2Dial and WoW respectively. Models are selected based on the best dev set EM score. The maximum length of dialogue context is 128. The maximum lengths of model input for each passage are 512 and 384 for Doc2Dial and WoW respectively, due to the larger variation in passage length in Doc2Dial. During training, we feed up to 20 passages into the models without posterior regularization. Otherwise the number of passages is reduced to 8 or 10 depending on the memory consumption. For inference, we feed up to 20 passages into the model for all experiment settings. Each training process is run on 2 NVIDIA Quadro Q6000 GPUs. It takes about 18 and 10 hours to train with or without posterior regularization for both datasets. The inference time takes less than 1 minutes per 1000 examples for all experiment settings on 2 GPUs. All our models based on uncased BERT base model contains between 110 to 115 million parameters. 

\section{Dataset Details}
\label{sec:appendix_data}
We follow all original preprocessing and evaluation scripts for data processing of the following datasets.\footnote{Doc2Dial: \url{https://github.com/doc2dial/sharedtask-dialdoc2021}; WoW: \url{https://github.com/facebookresearch/ParlAI/tree/master/parlai/tasks/wizard_of_wikipedia}} These original data preprocessing scripts contain the step of downloading data.

\paragraph{Doc2Dial} \cite{feng-etal-2020-doc2dial} contains about 4.8k English goal-oriented dialogues in 4 social-welfare domains, with an additional Covid-19 domain in the blind held-out test set. Each dialogue has an average of 14 turns, grounded on a long document with more than 1k tokens on average. Each user or agent turn is grounded in a sequence of knowledge spans as annotated in the dataset. In terms of the number of agent turns, there are about 20k / 4k examples in the train / dev set. The blind held-out test set contains 800 examples.

\paragraph{WoW} \cite{dinan2018wizard} contains over 20k English social chat conversations with an average of 9 turns on over 1k open-domain topics. For each agent turn, the agent (i.e., wizard) chose one or no grounding sentence from on average 7 Wikipedia passages retrieved by a pre-defined retriever based on the dialogue history for composing the response. Each passage contains 10 sentences. The original data has its dev/test set split to two subsets, which contain conversations about topics seen or unseen in training. It contains about 18k dialogues for training, 2k dialogues for validation and
2k dialogues for test. The test set is split into two subsets, Test Seen and Test Unseen. Test Seen contains 965 dialogues on the topics overlapped with the training set, while Test Unseen contains
968 dialogues on the topics never seen before in training and validation set.

\end{document}